\documentclass{article}

\usepackage[preprint]{nips_2018}

\usepackage[utf8]{inputenc} 
\usepackage[T1]{fontenc}    
\usepackage{hyperref}       
\usepackage{url}            
\usepackage{booktabs}       
\usepackage{amsfonts}       
\usepackage{nicefrac}       
\usepackage{microtype}      
\usepackage{adjustbox}
\usepackage{booktabs,siunitx}
\usepackage{multirow}
\usepackage{siunitx}
\usepackage[toc,page]{appendix}
\usepackage[style=numeric,backend=biber]{biblatex}
\bibliography{bibliography}
\usepackage{graphicx}
\usepackage{caption}
\usepackage{subcaption}
\usepackage[compact]{titlesec}
\usepackage{fullpage, verbatim,mathtools}
\usepackage{comment}
\usepackage{hyperref}
 \usepackage{catchfile}
\usepackage[toc,page]{appendix}

\title{Can We Achieve More with Less? \\ Exploring Data Augmentation for Toxic Comment Classification}

\author{
  Chetanya Rastogi \\
  Department of Computer Science\\
  Stanford University\\
  \And
   Nikka Mofid    \\
  Department of Electrical Engineering\\
  Stanford University\\
  \And
  Fang-I Hsiao\\
  Department of Electrical Engineering\\
  Stanford University\\
}

\begin{document}

\maketitle

\begin{abstract}
This paper tackles one of the greatest limitations in Machine Learning: Data Scarcity. Specifically, we explore whether high accuracy classifiers can be built from small datasets, utilizing a combination of data augmentation techniques and machine learning algorithms. In this paper, we experiment with Easy Data Augmentation (EDA) and Backtranslation, as well as with three popular learning algorithms, Logistic Regression, Support Vector Machine (SVM), and Bidirectional Long Short-Term Memory Network (Bi-LSTM). For our experimentation, we utilize the Wikipedia Toxic Comments dataset so that in the process of exploring the benefits of data augmentation, we can develop a model to detect and classify toxic speech in comments to help fight back against cyberbullying and online harassment. Ultimately, we found that data augmentation techniques can be used to significantly boost the performance of classifiers and are an excellent strategy to combat lack of data in NLP problems.
\end{abstract}

\section{Introduction}
Data, Hardware and Algorithms are often described as the three guiding pillars of machine learning.
While researchers have made huge advancements towards the latter two pillars, developing
multiple efficient, state-of-the-art learning algorithms \cite{yuan2019object,xie2019self, edunov2018understanding, provilkov2019bpe, vaswani2017attention, saon2017english, han2019state} as well as breaking the barriers of hardware computing power with cutting edge developments in model parallelism \cite{shoeybi2019megatron}, solving machine learning's data bottleneck still remains a challenge.There are millions of interesting problems in the world but, without enough data, high accuracy classifiers to solve these problems can not be built. This can be especially frustrating for problems where the solution is within reach, but a lack of data prevents breakthroughs. In the converse, for problems where data is abundant, labeling can be an extremely grueling and time consuming process.

Thus, in this study, we attempt to tackle this data divide and explore if high accuracy classifiers can be built from small datasets using a combination of data augmentation techniques and machine learning methods. In the process, we develop a model to detect and classify toxic speech in comments to help web moderators fight back against online harassment and cyberbullying, and also protect data annotators from the psychological stress of having to label an extremely large, graphic dataset.

\section{Literature Review}
 In the recent years, deep learning has seen some great advancements with the advent of transfer learning \cite{DBLP:journals/corr/abs-1810-04805,Pennington14glove:global}, better architectures \cite{DBLP:journals/corr/VaswaniSPUJGKP17}, and improved language models \cite{DBLP:journals/corr/abs-1801-06146}. But for performing the basic task of classification, the size of "labelled" training data still dictates the performance of a model \cite{MoreData2,Moredata1}. For computer vision tasks, automatic data augmentation has been commonly used in to enhance scare or limited datasets for the purpose of machine learning. Specifically, techniques such as  image manipulation techniques such as geometric and color space transformation as well as image mixing \cite{Krizhevsky:2017:ICD:3098997.3065386} which introduce noise into the data are used to produce more data from the original dataset. Though augmented data may be of lower quality it has been shown that, algorithms can actually perform better, as long as useful information can be extracted by the model from the original dataset. 
 
 In natural language processing, the existence of strong local structure in the context of language which make it  difficult to come up with generalized rules for language transformation. Not every word has a synonym with which it can be replaced and even if a synonym does exist, the context may direct the meaning of a sentence in a completely orthogonal direction \cite{gennari2007context}. \cite{kobayashi2018contextual} tackles this problem of generating a replacement of a word by deriving the meaning from the adjoining context. Similarly \cite{dao2013alternate} proposes to choose a replacement based on it's vector representation in the latent space which in itself, is derived from the context \cite{mikolov2013distributed}. To retain the same semantic of a text, \cite{sennrich-etal-2016-improving} proposes a pipeline to translate the sentence into some intermediary language and back and shows a significant improvement in the task of machine translation. \cite{EDA} proposes a simple and efficient bag-of-words model to perform augmentation and propose three new operations for text augmentation.
 
 In this study we aim to explore the data augmentation techniques for natural language processing proposed in the related works, Easy Data Augmentation  \cite{EDA} and Back Translation \cite{sennrich-etal-2016-improving}, in tandem with different machine learning methodologies. Apart from analysing both the approaches independently, we also perform experiments to interpret and explain the reasons of each technique's success and carry out comparative analysis with respect to different machine learning algorithms. We also show that the choice of data augmentation depends crucially on the hypothesis space of the underlying model and the gains diminish with an increase in the model's expressive capabilities.

\section{Infrastructure}

\subsection{Dataset}
  Our dataset is the "Wikipedia Toxic Comments" dataset  which we obtained from Kaggle through the Kaggle Comment Classification Challenge \cite{data}. The data contains a list of $\sim$158k Wikipedia comments and six binary labels for the kind of hate speech each comment qualifies as. For our task, we take any kind of toxicity as a positive class and rest of the comments as the negative class. Before training with this dataset, we preformed some preprocessing to clean the data and also to produce simplified labeling as "Toxic" (1) or "Non-toxic" (0). Our preprocessing script cleans the data by removing any special chars from comments and correcting spelling of key words. We utilize the final, preprocessed dataset as the data source for our experimentation.

\subsection{Training Structure}
 We split the data from our pre-processed dataset into train and test and keep the test set aside, as it will only be used for evaluation purposes. To evaluate the performance of data augmentation, we further sample 5\% of the data from the train set which serves as the small training set for our baseline (without data augmentation) and the entire train set serves as the oracle. In addition, we run our data augmentation algorithms on this small training set in order to evaluate the performance of the techniques when combined with different learning algorithms.

\subsection{Experimental Setup}
Our experimental pipeline is as follows. The raw text is first preprocessed to remove special characters such as @, \$, ?, etc and is converted to lower case to reduce the vocabulary size. Since many comments try to hide the toxicity by putting special characters or by changing the spelling we also normalize the text using certain regular expressions.

After the preprocessing is done, we split the dataset into train and test, and further sample 5\% of the train set as the "small training set" which we run through our different machine learning techniques, including logistic regression, SVM, and Bi-LSTM, for our baseline. We run our different data augmentation algorithms, specifically EDA and Backtranslation, on our "small training set" and then use the different machine learning methods mentioned above on these augmented datasets to see which combination works best. 

For EDA, we experiment with multiple values of $\alpha$ and found that at a value of 0.1-0.15 we get optimal results. As described in section \ref{EDA:implement}, we make use of \textit{nltk} library for finding stopwords and synonyms. We generate 9 augmented sentences per training sentence thus, increasing the size of the training set by 10x. For backtranslation, we use Spanish as our intermediary language and generate one augmented sentence per sentence in the training set thus, increasing the training set by 2x. 
    
For implementing the classification models, \textit{scikit-learn} is used to train LR and SVM models while the Bi-LSTM model is trained using \textit{keras} with tensorflow backend.  LR and SVM make use of TF-IDF feature vectors while the Bi-LSTM model uses Glove \cite{Pennington14glove:global} embedding layer. We will be going into detail about our learning and data augmentation algorithms in Section 5 and Section 6. The following diagram depicts the flow of information through our toxicity classification pipeline:

    \begin{figure}[h!]
        \centering
        \includegraphics[scale=0.75]{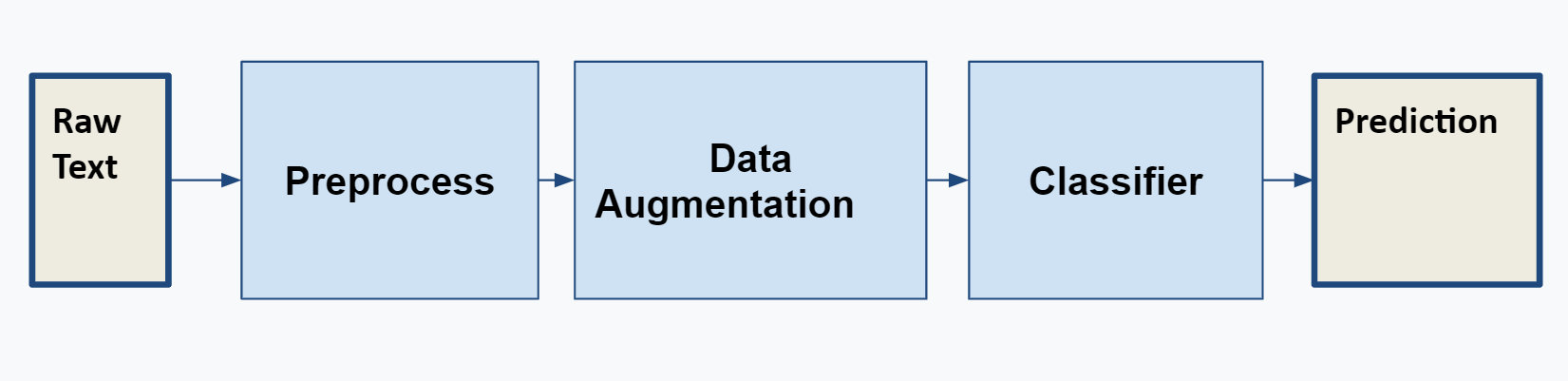}
        \caption{Toxicity classification pipeline}
        \label{fig:Pipeline}
     \end{figure}

\subsection{Evaluation Metrics}
It is important to note that our dataset is highly imbalances, with only 10\% of the comments classified as toxic. As a result, we will use the following metrics to evaluate our method:  Recall and F1 score. This is because these metrics take into account false positive and negatives, which are very crucial in such an uneven dataset.

 \section{Learning Algorithms} \label{learning_algos} 
    We will be using three learning algorithms of varying complexity as our classifiers. We compare the effect of our data augmentation algorithms on these learning methods.
    
    \subsection{Logistic Regression (LR)}
    Logistic regression is a simple learning algorithm used for classification that assigns observations to a discrete set of classes. It is different than linear regression in that it uses the logistic sigmoid function to return probability value which can be mapped to two or more discrete classes \cite{LR}. We chose to use logistic regression as one of the learning algorithms to analyze the performance of data augmentation because it is simple and very commonly used for classification tasks like this one so the performance results are very relevant to those undertaking similar research with small or limited datasets. 
    
    \subsection{Support Vector Machines (SVM)}
    SVM is a discriminative classifier which given labeled data, finds an optimal hyperplane in an N-dimensional space that uniquely categorizes the data points. There are many hyperplanes or "decision boundaries" that can be chosen when splitting the data into classes and an SVM works by finding the plane that has the maximum margin or distance between data points of both classes. The points that are closest to the hyperplane are support vectors and they are used by the SVM to maximize the margin of the classifier  \cite{SVM}. We chose to use an SVM as one of the learning algorithms to analyze the performance of data augmentation because it is more expressive than logistic regression and highly interpretable in contrast to a neural network. Due to its wide use for such problems and a slight increase in complexity, we thought that an SVM would provide an excellent performance comparison to logistic regression and, as we will see in the next section, the more complex bidirectional LSTM. \vspace{0.5em}
    
    \subsection{Bidirectional LSTM (Bi-LSTM)}
    LSTM is a common deep learning model for natural language processing tasks. It is a recurrent neural network with a more complicated architecture. We use bidirectional LSTM which reads the text forward and backward and then combines these two features to make the prediction \cite{LSTM}. We chose to use bidirectional LSTM as one of our learning algorithms to analyze the performance of data augmentation because unlike SVM and Logistic Regression (LR), LSTMs can capture the sequence of tokens which adds an extra layer of information not captured by the bag-of-words model of SVM and LR. Thus, it allows us to see the performance of data augmentation on a more complicated and precise learning algorithm which may be applied to such a classification task.
    
    \section{Data Augmentation Algorithms}
    We have implemented different data augmentation techniques and compare their effect in terms of the performance boost on different types of learning methods(described in section \ref{learning_algos}).We chose to focus on Easy Data Augmentation and Backtranslation because they are two of the most popular and effective methods of data augmentation \cite{EDA}.

    \subsection{Easy Data Augmentation(EDA)} 
    \subsubsection{Algorithm \cite{EDA}}
    EDA makes use of four extremely simple operations to generate new augmented sentences given a basis sentence. The algorithm takes in one sentence from the training set and performs one of the following operations chosen at random: 
    \begin{enumerate}
        \item \textbf{Synonym Replacement(SR):} Randomly choose \textit{n} tokens from the sentences that are not stop words. Each chosen token is then replaced by one of its synonym chosen at random.
        \item \textbf{Random Swap(RS):} Randomly choose \textit{n} pairs of tokens from the sentence and swap their positions.
        \item \textbf{Random Insertion(RI):} This step is a slight advancement of SR. Like SR, again choose \textit{n} tokens at random from the sentence which are not stop words but instead of replacing them with their 
        respective synonyms, insert the synonym (one for each token) at a random position in the sentence. 
        \item \textbf{Random Deletion(RD):} For each token in the sentence, determine whether to delete it or not with a probability \textit{p}. 
    \end{enumerate}
    Each of the above four operations has an intuition behind it that works on tackling different problems that are mostly encountered while working on a small dataset. SR helps in maintaining the same syntactic and semantic meaning \cite{Zhang:2015:CCN:2969239.2969312} as the original sentence while providing an opportunity to the model to "learn" about new words that might not be present in the original dataset. RI and RS maintain all the original tokens in the sentence but introduce perturbations that helps in regularization and as a result, the model generalizes well when it comes across unknown patterns. RD helps in reducing model overfitting that is generally the case with a small dataset. By deleting words at random, it makes sure that the model is not "memorizing" particular patterns that it sees in the small dataset and forces it to explore other features.
   
    \subsubsection{Implementation Details}\label{EDA:implement}
    The aim of any data augmentation technique is to produce new data points while preserving the original class label. We implement EDA in such a way that the amount of augmentation a sentence undergoes(number of tokens inserted, swapped, deleted, or replaced) is proportional to the length of the sentence. This is because long sentences are more robust to noise than short sentences because it's very easy to destroy a structure of a sentence having fewer tokens. A parameter ($\alpha$) dictates how many tokens(\textit{n}) will undergo RS,SR, RI(\textit{n=$\alpha$l, l} being the length of the sentence) and serves as a hyperparameter that needs to be tuned. For the sake of consistency, the probability with which a token is deleted(RD) is also set to $\alpha$.
    
    We define four functions, one corresponding to each operation, which are called uniformly at random to generate an augmented instance of a given instance. We also provide the flexibility of how many augmented sentences does a user want to generate per sentence. For SR, we make use of python's \textit{nltk} library for defining stop words and finding synonyms while the rest of the functions are fairly straightforward and doesn't require any external library support. Examples of augmented sentences are shown in Table \ref{tab:EDA_examples}
    \begin{table}[h!]
    \centering 
    \begin{tabular}{p{0.4\linewidth}cp{0.4\linewidth}l}
    \toprule 
    \textbf{Original Sentence} & \textbf{Operation} & \textbf{Augmented Sentence} \\
    \midrule
    how can you \textbf{block} me when you're just an editor    & SR     & how can you \textbf{impede} me when youre just an editor  \\
    how can you \textbf{block} me when you're just an editor & RD  & how can you me when youre just an editor     \\
    how \textbf{can} you block me \textbf{when} you're just an editor & RS  & how \textbf{when} you block me \textbf{can} youre just an editor \\
    how can you block me when you're \textbf{just} an editor    & RI & how can you block \textbf{simply} me when youre just an editor \\
     \bottomrule
    \end{tabular}
    \caption{Augmented Sentences using EDA Operations}\label{tab:EDA_examples}
    \end{table}
    
    \subsection{Backtranslation}
    \subsubsection{Algorithm}
    The idea of backtranslation was introduced to generate parallel sentences in two languages for improving the performance of Neural Machine Translation models \cite{sennrich-etal-2016-improving}. The key idea is to use the noise introduced by NMT models to augment training data. Since NMT models try to retain the same semantic meaning as the source language, the generated output is a rephrased version of the same sentence, thus providing new data points for the models to train on. 
    
    Backtranslation works by taking a sentence from the training set(source) and translating it to an intermediate language and then translating the generated output back to the source language using an NMT model. This process results in the introduction of noise at both the translation stages and we end up with a new sentence having similar semantic meaning as the original one. The pipeline for backtranslation is as follows: 
    
      \begin{figure}[h!]
        \centering
        \includegraphics[scale=0.75]{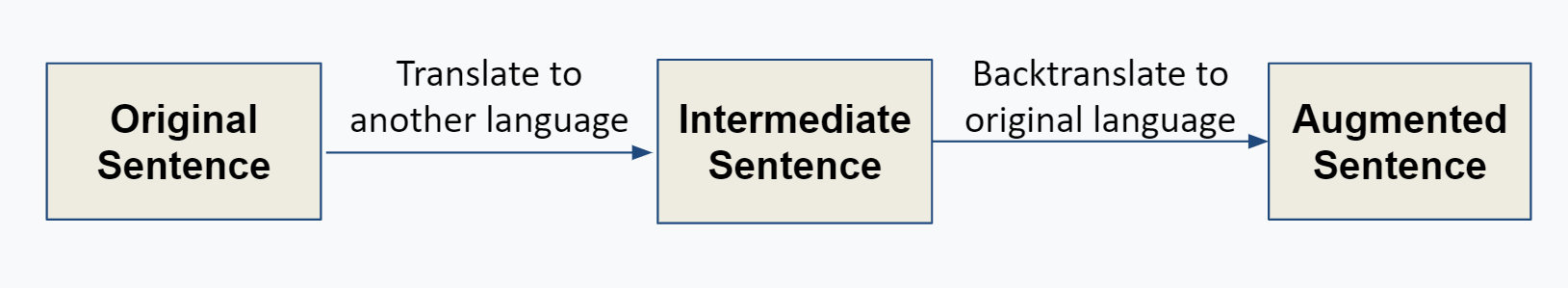}
        \caption{Backtranslation pipeline}
        \label{fig:Pipeline}
     \end{figure}

    \subsubsection{Implementation Details}
    We make use of the paid Google Translate API \cite{GoogleTranslate} for this task. A script was written that links with the user account on Google Cloud Platform and makes HTTPS requests to the API. The API supports more than 50 languages. We chose 4 languages viz., Spanish, French, Hindi and German as our intermediary languages where each language provides a different aspect in contrast to English. An example of such an augmented sentence is shown below in Table \ref{tab:BT_examples}
    \begin{table}[h!]
    \centering 
    \begin{tabular}{p{0.4\linewidth}cp{0.4\linewidth}l}
     \toprule
     \textbf{Original Sentence} & \textbf{Operation} & \textbf{Augmented Sentence} \\
     \midrule
    I too agree with your suggestion thanks for taking this on  & BT & I also agree with your suggestion thank you for taking this \\
    \bottomrule
    \end{tabular}
    \caption{Augmented Sentences using Backtranslation (BT)}\label{tab:BT_examples}
    \end{table}

\section{Results \& Discussion}
\begin{figure}[h!]
        \centering
        \includegraphics[scale=0.75]{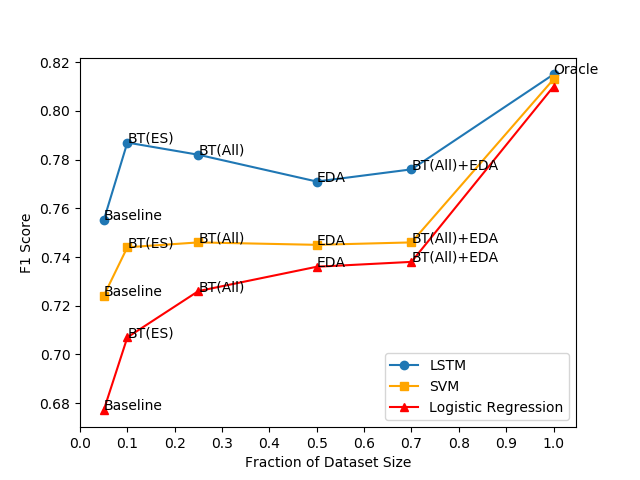}
        \caption{F1 Score vs Size of Dataset (in terms of Fraction of Full Dataset size)}
        \label{fig:Peformance}
     \end{figure}

    Tables \ref{table:F1} and \ref{table:Recall} as well as Figure \ref{fig:Peformance} detail our results for each machine learning and data augmentation algorithm combination.
    \begin{table}[h!]
    \centering 
    \begin{tabular}{lllc}
    \toprule
              &LR & SVM & Bi-LSTM    \\
    \midrule
    Baseline  & 0.6770    & 0.7240     & 0.7555  \\
    EDA      & 0.7360      & 0.7453         & 0.7712    \\
     Backtranslation (DE) & 0.7022   & 0.7363       & 0.7614   \\
    Backtranslation (FR) & 0.6941    &  0.7314      & 0.7823    \\
    Backtranslation (HI) & 0.7060   & 0.7417       & 0.7829 \\
    Backtranslation (ES) & 0.7079   & 0.7444      & 0.7872   \\
    Backtranslation (ALL) & 0.7264    & 0.7458     &0.7827  \\
    Backtranslation (ALL) + EDA  &0.7384  &0.7462    &0.7759   \\
    Oracle & 0.8010 & 0.8130 & 0.8155 \\
    \bottomrule
    \end{tabular}\caption{F1 Score Comparison}\label{table:F1}
    \end{table}

    \begin{table}[h!]
    \centering 
    \begin{tabular}{lllc}
    \toprule
              &LR & SVM & Bi-LSTM    \\
    \midrule
    Baseline  & 0.5270    & 0.5983     & 0.686  \\
    EDA     & 0.6290      & 0.6623    &0.7109     \\
    Backtranslation (DE) & 0.5658   &0.6262       &0.8092   \\
    Backtranslation (FR) & 0.5586    &0.6226      & 0.7695    \\
    Backtranslation (HI) & 0.5705    &0.6345      & 0.7423    \\
    Backtranslation (ES) &0.5764   &0.6410     &  0.7494 \\
    Backtranslation (ALL) & 0.6096  &0.6570     & 0.7683    \\
    Backtranslation (ALL) + EDA  &0.6404  &0.6724       &0.7518   \\
    Oracle & 0.7240 & 0.7393 & 0.795 \\
    \bottomrule
    \end{tabular}\caption{Recall Comparison}\label{table:Recall}
    \end{table}

As mentioned in Section 3, because of the highly imbalanced nature of our data set (only 10 percent of the comments are toxic) we use Recall and F1 score to evaluate our dataset. Clearly both the data augmentation techniques show improvement over the baseline with an average F1 score improvement of 3\% on EDA and 2.3\% on backtranslation over all the 4 methods. The same is true for Recall as the EDA shows an average improvement of 6.3\% while backtranslation gives an average boost of 6\%, showing that data augmentation can improve the performance of classifiers.

\subsection{Backtranslation vs EDA}
To understand these improvements, we plot the graphs of feature importances (Refer Appendix \ref{Plots_and_figures}) for all three classifiers. As evident from the figures, both EDA and backtranslation help in boosting the importance of the top features to the level obtained in the oracle(full dataset). Also simple classifiers such as LR and SVM receive a major boost from both the augmentation techniques whereas Bi-LSTM shows a greater improvement in performance with backtranslation than with EDA. This can be attributed to the fact that since EDA is more suitable for bag-of-words model as it treats every token independently and disregards the semantic structure whereas backtranslation retains the semantic structure and hence provides better augmentation for Bi-LSTM model. 

In turn, this analysis is supported by Figure \ref{fig:Peformance} as well as Table \ref{table:F1} and \ref{table:Recall} which show that the F1 and Recall scores for logistic regression and SVM received the largest boost from backtranslation combined with EDA, as they are able to reap the benefits from both augmentation techniques while in contrast Bi-LSTM received the largest boost from backtranslation on its own, due to the preservation of semantic sentence structure. 

\subsection{Effect of Different intermediary language for Backtranslation}
We experimented with 4 different languages to analyse if a language's morphology affects the data augmentation process. The languages we choose are diverse and are representative of different aspects that contrast with English. Spanish is chosen as it is the second most spoken language in the world \cite{spanish} and hence has a high probability that NMT models trained on English-Spanish would have higher performance due to availability of high-quality data. French is considered to share many similarities with English including the same alphabet and a number of true cognates \cite{French:English} while Hindi has a completely different etymology. German is used as a lot of work has been done in the English-German machine translation tasks which gives access to better quality models. 

\begin{table}[h!]
    \centering 
    \begin{tabular}{lc}
    \toprule
        \textbf{Dataset} & \textbf{Vocab Size} \\
    \midrule
    Baseline  & 29978   \\
    Backtranslation (DE) & 31676 \\
    Backtranslation (FR) & 32748 \\
    Backtranslation (HI) & 36975    \\
    Backtranslation (ES) & 33460 \\
    Backtranslation (ALL) & 41167    \\
    \bottomrule
    \end{tabular}\caption{Vocabulary size of baseline dataset in different languages }\label{table:vocab}
    \end{table}

Comparing between the four backtranslation languages, we received the best results with Spanish while Hindi comes in close second. We hypothesize that these languages had colloquial nuances which caused entirely new words to be introduced into the translated text teaching the algorithm to generalize while also maintaining the general semantics of the sentence. Table \ref{table:vocab} shows the size of vocabulary after the baseline dataset was augmented with the different languages.

\section{Error Analysis}
Looking at our feature importance graphs, we do see some error despite the generally good performance of our algorithm. For example, in the feature importance graphs for logistic regression in Figure \ref{fig:LR}, we see that for the Baseline+Backtranslation (ES), the word \textit{you} is classified with the same,albeit low, level of toxicity as extremely hurtful words like \textit{faggot} and \textit{bastard}. This could be because \textit{you} is often in the same sentences as a toxic word or toxic noun which leads the classifier to associate it with toxicity.  We see this again in the feature importance plots for the bi-LSTM, specifically for Figure \ref{fig:sub2_EDA} which represents Baseline+EDA. 

Another kind of error that creeps into the system results from words that have both toxic as well as non-toxic connotations associated with them. One such example is the word \textit{wasted}. Similarly to the previous example, \textit{wasted} could be misclassified as toxic because it often appears in sentences with toxic words, or because of its double meaning in the English language, where formally it means to "use or expend carelessly" while colloquially it is used to refer to someone "under the influence of alcohol or drugs". The double meaning of the word "wasted" shows how the context of a sentence can influence the meaning of the word and shows that our classifier may still need some work to deal with words with multiple meanings in order to disambiguate correctly. 

\section{Conclusion and Future Work}
Our results show that data augmentation and classifier selection go hand in hand, and data augmentation is an excellent and robust solution to combat data limitations in natural language processing problems. Augmentation techniques used by EDA are more favorable for algorithms that operate under a bag-of-words model while Backtranslation is suitable when the semantic structure of the text is also taken into account. We also try to explain how each of these techniques modulate the data and provide a subsequent gain in performance by providing interpretable insights in terms of the feature importance. 

For potential future work, we aim to explore if we can make use of the abundant unlabeled data to make more robust classifiers. The paradigm of positive-unlabeled learning and unsupervised learning can be another area which can be investigated further in this regard. Overall, our study has a great social impact in that we have not only implicitly created a system to detect and classify toxic comments but also explored data augmentation algorithms which can combat data limitations in machine learning research and also help annotators by decreasing the amount of data they might have to label. We are very excited for potential future work!

  \printbibliography
  
  \begin{appendices}
      \section{Plots and Figures}\label{Plots_and_figures}
\begin{figure}[h!]
\centering
\begin{subfigure}{.55\textwidth}
  \centering
  \includegraphics[width=\linewidth]{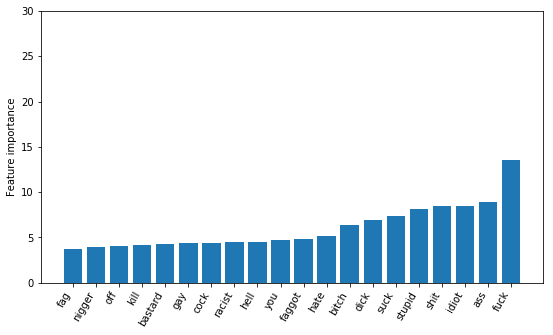}
  \caption{Baseline}
  \label{fig:sub1}
\end{subfigure}%
\begin{subfigure}{.55\textwidth}
  \centering
  \includegraphics[width=\linewidth]{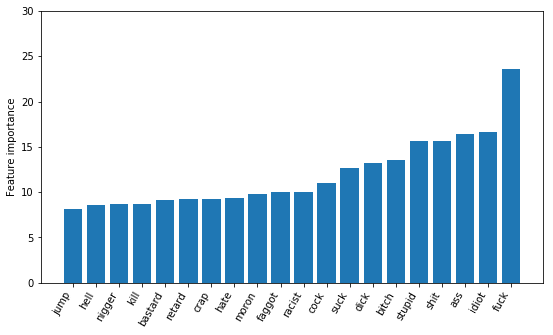}
  \caption{Baseline+EDA}
  \label{fig:sub2}
\end{subfigure}
\begin{subfigure}{.55\textwidth}
  \centering
  \includegraphics[width=\linewidth]{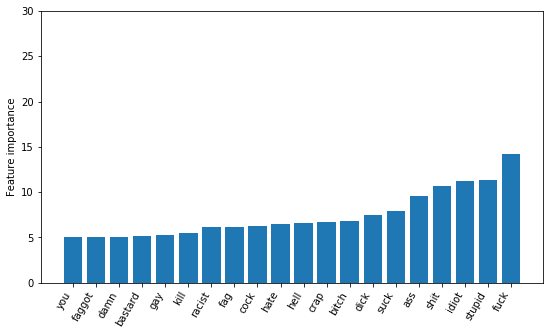}
  \caption{Baseline+Backtranslation (ES)}
  \label{fig:sub1}
\end{subfigure}%
\begin{subfigure}{.55\textwidth}
  \centering
  \includegraphics[width=\linewidth]{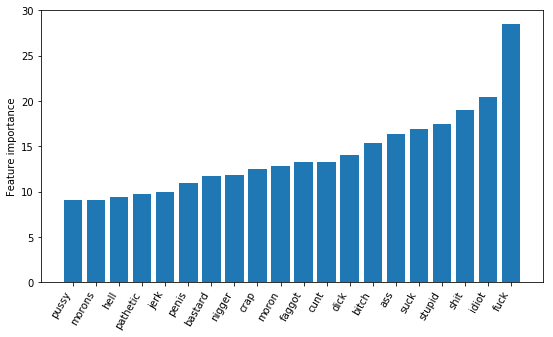}
  \caption{Oracle}
  \label{fig:sub1}
\end{subfigure}%
\caption{Feature importance for Logistic Regression}
\label{fig:LR}
\end{figure}

\begin{figure}[h]
\centering
\begin{subfigure}{.55\textwidth}
  \centering
  \includegraphics[width=\linewidth]{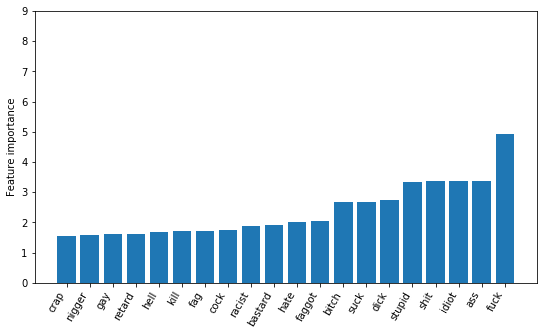}
  \caption{Baseline}
  \label{fig:sub1}
\end{subfigure}%
\begin{subfigure}{.55\textwidth}
  \centering
  \includegraphics[width=\linewidth]{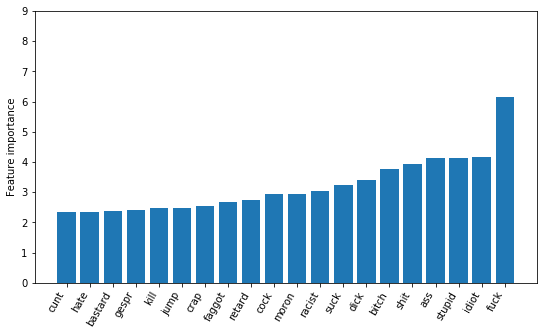}
  \caption{Baseline +EDA}
  \label{fig:sub2}
\end{subfigure}
\begin{subfigure}{.55\textwidth}
  \centering
  \includegraphics[width=\linewidth]{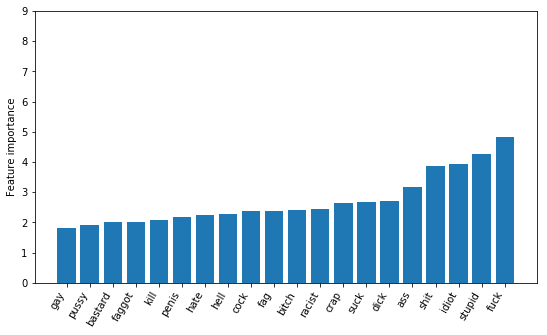}
  \caption{Baseline+Backtranslation (ES)}
  \label{fig:sub1}
\end{subfigure}%
\begin{subfigure}{.55\textwidth}
  \centering
  \includegraphics[width=\linewidth]{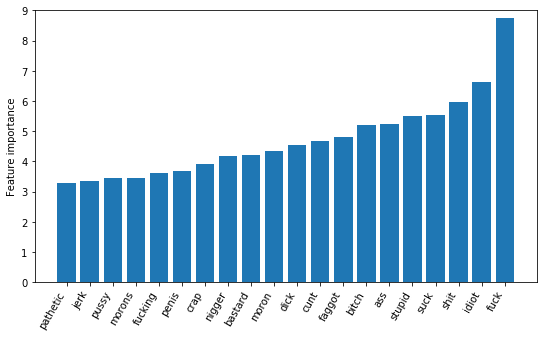}
  \caption{Oracle}
  \label{fig:sub1}
\end{subfigure}%
\caption{Feature importance for SVM}
\label{fig:SVM}
\end{figure}

\begin{figure}[h]
\centering
\begin{subfigure}{0.67\textwidth}
  \centering
  \includegraphics[width=\linewidth]{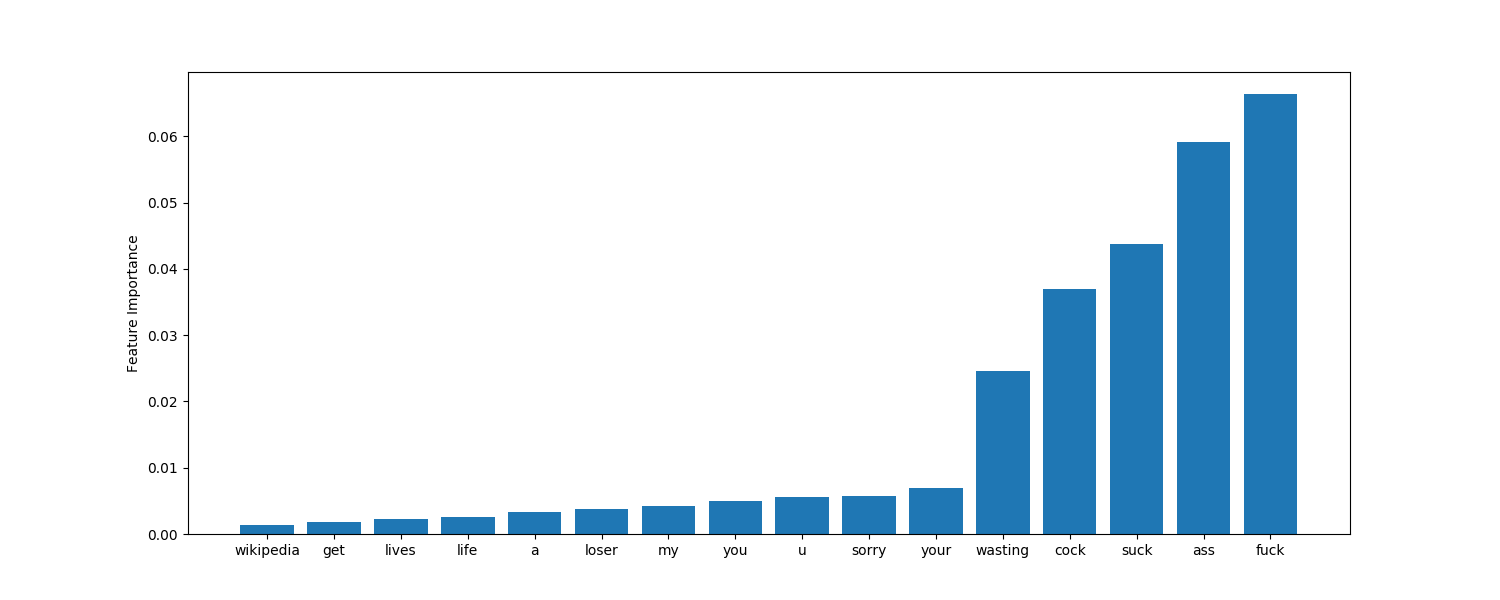}
  \caption{Baseline}
  \label{fig:sub1}
\end{subfigure}
\begin{subfigure}{0.67\textwidth}
  \centering
  \includegraphics[width=\linewidth]{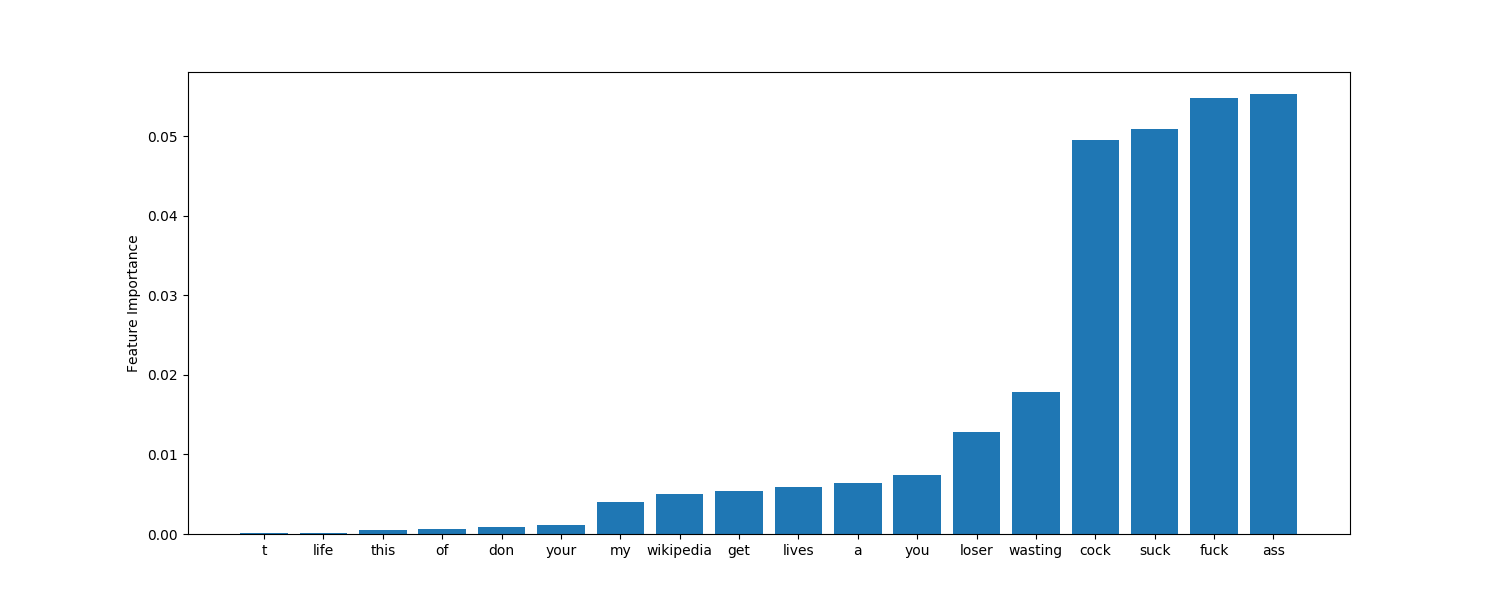}
  \caption{Baseline+EDA}
  \label{fig:sub2_EDA}
\end{subfigure}
\begin{subfigure}{0.67\textwidth}
  \centering
  \includegraphics[width=\linewidth]{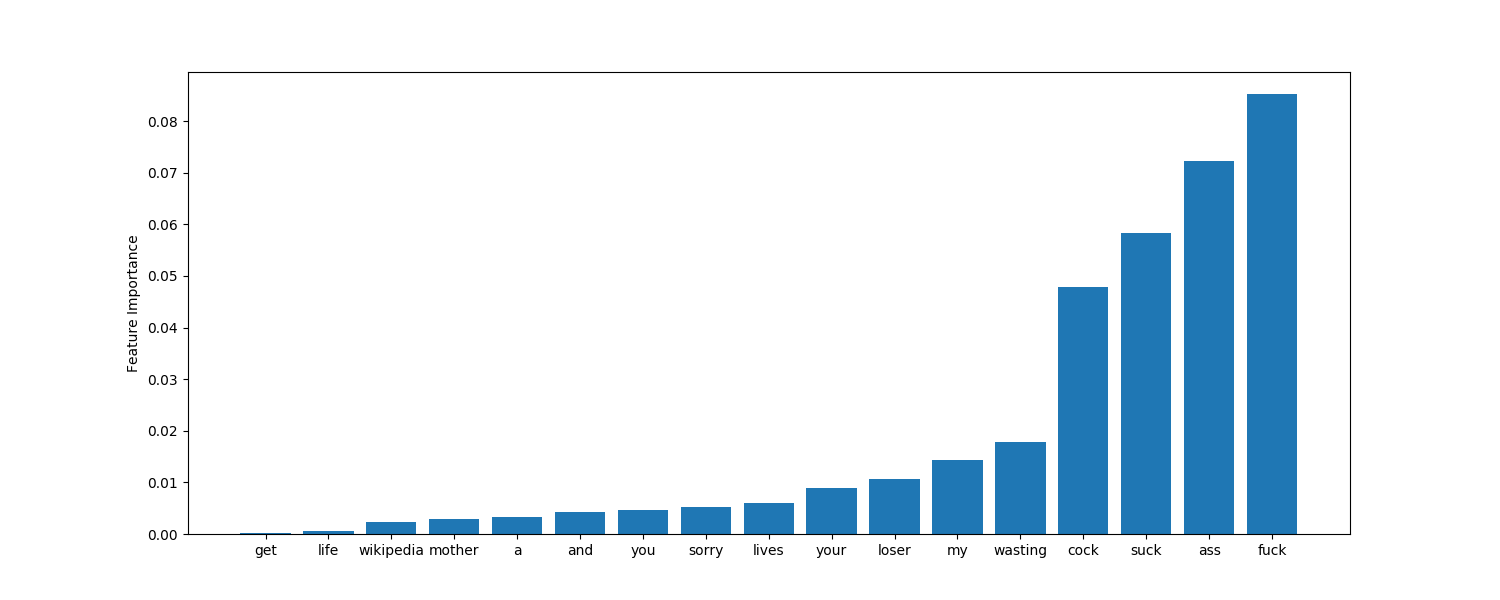}
  \caption{Baseline+Backtranslation(ES)}
  \label{fig:sub1}
\end{subfigure} 
\begin{subfigure}{0.67\textwidth}
  \centering
  \includegraphics[width=\linewidth]{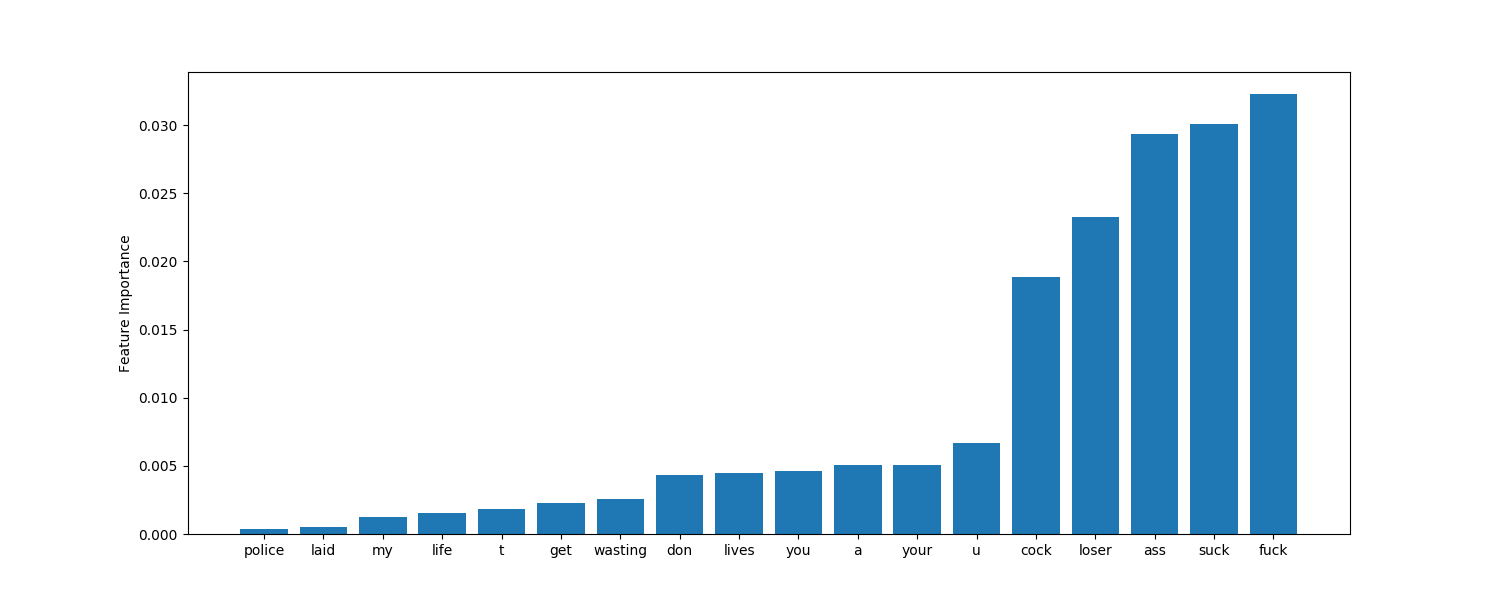}
  \caption{Oracle}
  \label{fig:sub1}
\end{subfigure}%
\caption{Feature importance for Bi-LSTM}
\label{fig:LSTM}
\end{figure}

    \end{appendices}
 
\end{document}